\documentclass[letterpaper, 10 pt, conference]{ieeeconf}
\IEEEoverridecommandlockouts
\usepackage[nocompress]{cite}
\usepackage{amsmath,amssymb,amsfonts}
\usepackage{algorithmic}
\usepackage{graphicx}
\usepackage{textcomp}
\usepackage{xcolor}
\usepackage{listings}
\usepackage{comment}
\usepackage{tikz}

\def\BibTeX{{\rm B\kern-.05em{\sc i\kern-.025em b}\kern-.08em
    T\kern-.1667em\lower.7ex\hbox{E}\kern-.125emX}}

\RequirePackage[                    
    strict=true,                    
    style=english                    
]{csquotes}

\usepackage{hyperref}
\usepackage{booktabs}

\newcommand\copyrightnotice[1]{
    \begin{tikzpicture}[remember picture,overlay]
    \node[anchor=south,yshift=10pt] at (current page.south) {\fbox{\parbox{\dimexpr\textwidth-\fboxsep-\fboxrule\relax}{#1}}};
    \end{tikzpicture}
}

\begin{document}

\title{\LARGE \bf Explain yourself!\\ Effects of Explanations in Human-Robot Interaction}

\author{Jakob Ambsdorf$^{*}$, Alina Munir, Yiyao Wei, Klaas Degkwitz, Harm Matthias Harms, Susanne Stannek,\\ Kyra Ahrens, Dennis Becker, Erik Strahl, Tom Weber, Stefan Wermter
\thanks{$^{*}$Corresponding author: jakob.ambsdorf@uni-hamburg.de}
\thanks{Acknowledgment: The authors gratefully acknowledge partial support from
the German Research Foundation DFG under project
CML (TRR 169).}
\thanks{Authors are with the Knowledge Technology Group (WTM), Department of Informatics, Universit\"at Hamburg, Vogt-K\"olln-Stra\ss e 30, \mbox{Hamburg D-22527}, Germany}
\thanks{Many thanks to Anirban Bhowmick, Jovana Bunjevac, Syed Hasan and Vincent Rolfs for their work and dedication in the HRI project.}
}

\maketitle


\begin{abstract}

Recent developments in explainable artificial intelligence promise the potential to transform human-robot interaction:
Explanations of robot decisions could affect user perceptions, justify their reliability, and increase trust.
However, the effects on human perceptions of robots that explain their decisions have not been studied thoroughly. 
To analyze the effect of explainable robots, we conduct a study in which two simulated robots play a competitive board game. While one robot explains its moves, the other robot only announces them.
Providing explanations for its actions was not sufficient to change the perceived competence, intelligence, likeability or safety ratings of the robot.
However, the results show that the robot that explains its moves is perceived as more lively and human-like.
This study demonstrates the need for and potential of explainable human-robot interaction and the wider assessment of its effects as a novel research direction.

\end{abstract}


\section{Introduction}
\label{sec:introduction}

\copyrightnotice{\small{\copyright 2022 IEEE.  Personal use of this material is permitted.  Permission from IEEE must be obtained for all other uses, in any current or future media, including reprinting/republishing this material for advertising or promotional purposes, creating new collective works, for resale or redistribution to servers or lists, or reuse of any copyrighted component of this work in other works}}
Explainable Artificial Intelligence (XAI)~\cite{arrieta2020explainable} promises to provide humanly understandable explanations about the actions, recommendations, and underlying causes of Artificial Intelligence (AI) techniques.
Explanations of an algorithm's decisions can increase transparency and build trust in users of artificial intelligence and robotic systems~\cite{Herlocker2000,Biran2017}. In Human-Robot Interaction (HRI), robots that explain their actions can increase users' confidence, trust, and reduce safety risks in human-robot cooperative tasks~\cite{Weidemann2021,Wang2016}, whereas miscommunication can create confusion and mistrust~\cite{Nomura2011}.

While the field of XAI has the potential to enhance human-robot interaction, there is still a lack of research on the effects and perception of robots that explain their actions to users. Specifically, due to the currently limited functionality and challenges to generate humanly understandable explanations from deep neural networks, the research on the interaction of humans and explainable robots is in its early stages of development~\cite{Anjomshoae2019}.
Despite the increasing attention on the application of XAI and its benefits, only a small amount of user studies have been conducted~\cite{Ehsan2019,Anjomshoae2019}. A considerable part of research in the field focuses on the design of theoretical frameworks for human-robot interaction studies that leverage XAI~\cite{Ciatto2020,Sanneman2020}.

To measure the effect of XAI in human-robot interaction, we conduct a study in which two simulated robots compete in a board game. 
While one of the robots explains the reasoning behind each of its game-play decisions, the other is constrained to only announcing each move without any explanations.
For the game scenario, the game Ultimate Tic-tac-toe was selected. The game is similar in rules to regular Tic-tac-toe but is less well-known, more challenging to play, and game states are difficult to analyze~\cite{George2016}.
For measuring the effect of a robot explaining its actions, the Godspeed series~\cite{bartneckHRIMetrics2009}, perceived competence, and participants' predictions about which robot will win are assessed. While the Godspeed questionnaire measures the general robot perception, perceived competence measures an underlying dimension of forming trust.

\begin{figure}[tbp]
    \centering
    \includegraphics[width=0.48\textwidth]{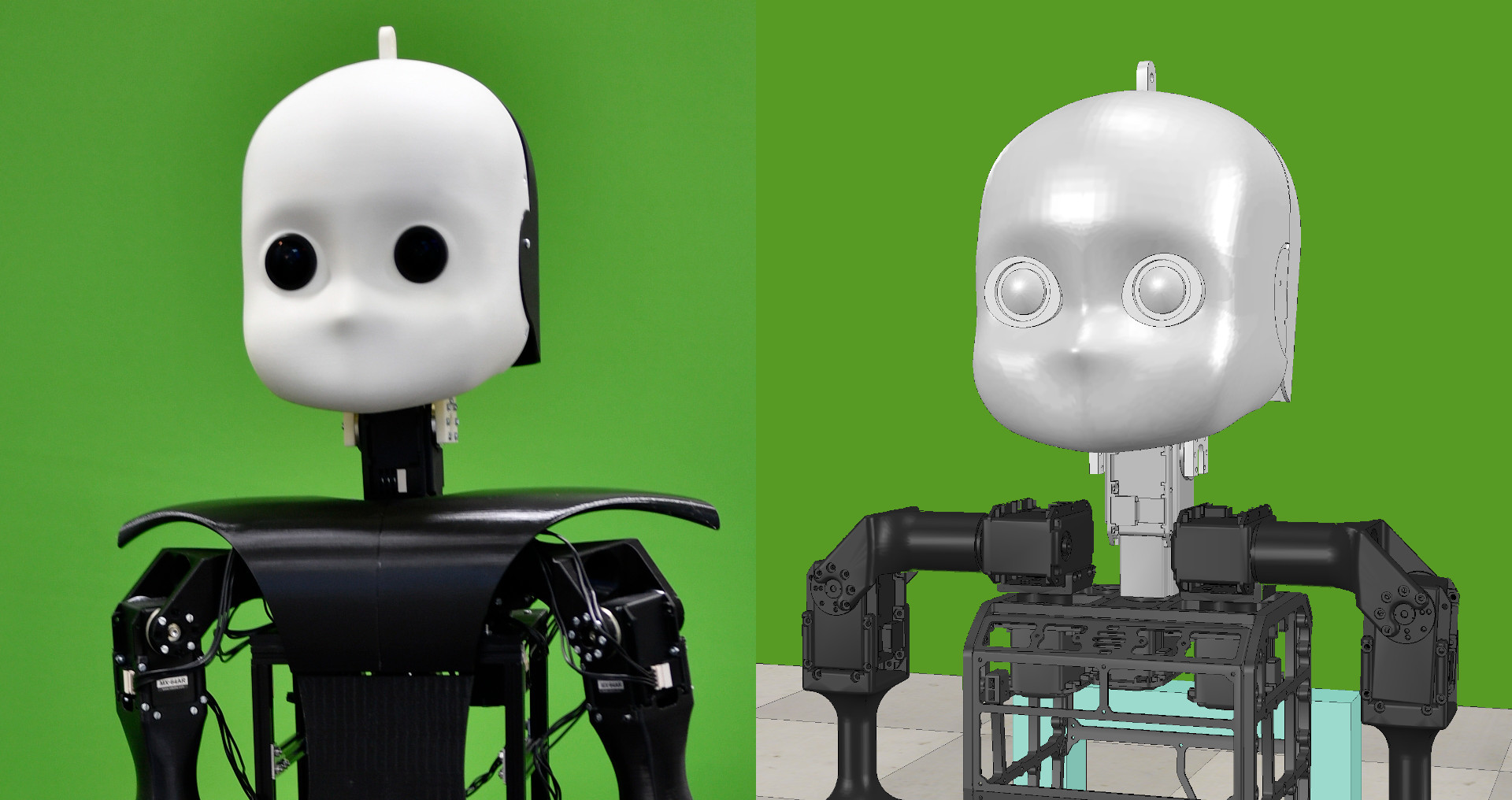}
    \caption{The real NICO robot on the left and its simulated counterpart. This study was conducted using two simulated robots.}
    \label{fig:nico}
\end{figure}

The technical components of the study are implemented in robot simulation software and can be transferred to real robots (see Figure~\ref{fig:nico}), enabling the study to be conducted in a lab experiment in the future. Due to the COVID-19 pandemic, the study was conducted entirely online.

\label{introduction}

\section{Related Work}
\label{sec:relatedwork}

The research on explainable artificial intelligence reaches back to rule-based expert systems in the 1970’s~\cite{Shortliffe1975,Confalonieri2021}. With the introduction of increasingly performant models, the importance of providing explanations and justifications for their actions was stressed early on ~\cite{Swartout1983,Chandrasekaran1991}.
The recent success of deep learning~\cite{Lecun2015} and the ubiquitous use of such artificial intelligence systems, which are largely considered to be black-boxes, has intensified the discussion and need for explainable AI considerably.
Thereupon, explainable artificial systems have been declared an economic priority, which causes governmental regulations and technical requirements~\cite{Bundy2017,Veiber2020}.
For example, citizens of the European Union, who are affected by automated decision-making systems, have the right to an explanation of that decision~\cite{Hamon2020}. Such regulations enforce that AI systems are transparent, explainable, and accountable for safety-critical applications~\cite{Preece2018,Goodman2017,Wachter2017}.

Consequently, there has been a surge in research of explainable systems under various names such as comprehensible, understandable, and explainable AI~\cite{adadi2018peeking}. 
Various attempts at taxonomizing the methods for explaining neural networks and other machine learning algorithms have been proposed recently. The properties of these methods, such as their dependence on the employed model, the degree of faithfulness to the prediction process and their means of communication, vary considerably among them. We direct the reader to corresponding surveys and taxonomies for a comprehensive overview~\cite{adadi2018peeking, arrieta2020explainable, dovsilovic2018explainable, samek2019towards}.
The majority of the methods that have been proposed generate explanations that are understandable only for experts, rather then potential users~\cite{miller2017explainable}. On the other hand, in the field of autonomous robotics~\cite{Sakai2021,Hussain2021} and in human-robot interaction~\cite{Setchi2020,Anjomshoae2019}, explaining actions to a user rather than an expert is particularly important. This is especially pronounced in human-robot cooperative tasks~\cite{Lemaignan2017}. In this context, explanations provide new information to the user, thereby assisting in understanding the reasoning behind actions, as well as providing an intuition about the robot's capability~\cite{Lombrozo2006, Sanneman2020}.

Trust has been identified as the foundation of human-robot interaction for collaborative and cooperative tasks~\cite{Hancock2011,Khavas2020}.
Explanations can be considered a human-like attribute, and research has shown that humans expect explanations in conjunction with a robot's actions~\cite{Han2021}.
An accurate understanding of the robot's intent, which is derived by explanations of the robot's decision-making, can consequently increase trust~\cite{Wang2016}. Additionally, research suggests that the robot's display of human-like attributes can lead to fostering trust as well~\cite{Natarajan2020}.
However, trust can be lost quickly in the event of inconsistent actions~\cite{Sebo2019} or robot failure~\cite{Desai2013}.

One dimension of human-robot trust is the robot's competence to achieve the user's desired goal~\cite{Park2020}.
The perception of a robot as competent is a requirement for the development of trust~\cite{Christoforakos2021} and changes the social human-robot relationship~\cite{Tian2021}. The robot's competence has been suggested as one of the main factors determining the preference of one robot over another in HRI scenarios~\cite{scheunemann2020warmth}. 
Concurrently, users tend to over- or underestimate a robot's competence based on their perception and observed robot behavior~\cite{Cha2015}.
Despite the significance of perceived competence for successful human-robot interaction and its link to trust, there has been, to our knowledge, little work directly assessing how explanations can impact perceived competence.

To assess the perception of robots, remote~\cite{Canning2014,Schermerhorn2011,Kiesler2008} and video-based studies~\cite{Walters2011,Lehmann2016,Syrdal2010} have been successfully conducted.
Prior work found that video-based HRI studies generally achieve comparable results with their live counterparts~\cite{Woods2006}. However, it has been shown that certain aspects, such as empathy for a robot and willingness to cooperate, can be greater in physical interaction with a robot as opposed to interacting with the robot remotely~\cite{Bainbridge2011,Seo2015}.
Due to the COVID-19 pandemic, researchers further explored the possibilities of remote HRI studies and noticed that this type of experiment can increase the participants' effort and their frustration~\cite{Gittens2021}.
In contrast to live HRI studies, video-based studies have the advantage of reaching more participants and consistent experiment conditions among participants~\cite{Jung2021}.
\label{relatedwork}
\section{Methods}
\label{sec:methods}

\subsection{Participants}
Data for the study were collected from February 19th, 2021, to March 3rd, 2021.
For the study evaluation, only the data of participants that left a remark about the study's objective and required at least 20 minutes to complete the study are considered.
This left a total of 92 participants for evaluation. Detailed demographic information is shown in Table~\ref{table:demographics}.

\begin{table}[htbp]
  \centering
  \caption{Demographic information about the participants}
  \label{table:demographics}
  \begin{tabular}{l r r}
      \toprule
      \textbf{\emph{Gender}} & \textbf{Frequency} &\textbf{\%}\\
      \midrule
      Male                      & 42    & 45.66 \\  
      Female                    & 43    & 46.74 \\
      Other                     & 7    & 7.60 \\
      \midrule
      \textbf{\emph{Age}} & \textbf{Frequency} &\textbf{\%}\\
      \midrule
      $< 20$                & 4     & 4.35 \\
      $20-25$               & 40    & 43.48 \\
      $26-30$               & 26    & 28.26 \\
      $31-35$               & 9    & 9.78 \\
      $36-40$               & 2     & 2.17 \\
      $> 40$                & 11    & 11.96 \\     
      \midrule
      \textbf{\emph{Nationality}} & \textbf{Frequency} &\textbf{\%}\\
      \midrule
      Asia          & 5    & 5.44 \\
      Germany       & 69    & 75.00 \\
      Rest of Europe & 7   & 7.61 \\
      Rest of the world & 2 & 2.17 \\
      No information     &  9  &   9.78 \\
      \midrule
      \textbf{\emph{Occupation}} & \textbf{Frequency} & \textbf{\%}\\
      \midrule
      Full or part-time employed    & 30      & 32.61 \\
      Not employed         & 2       & 2.17 \\
      Student               & 54      & 58.70 \\
      Retired               & 2       & 2.17 \\
      Other                 & 4       & 4.35 \\     
      \midrule
      \textbf{\emph{Profession}} & \textbf{Frequency} & \textbf{\%}\\
      \midrule
      Computer science relation      & 48    & 52.17 \\
      Other                 & 44    & 47.83 \\
      \midrule
      \textbf{\emph{UTTT Experience}} & \textbf{Frequency} & \textbf{\%}\\
      \midrule
      Experience with UTTT      & 8    & 8.70 \\
      No experience with UTTT    & 84    & 91.30 \\
      
    \bottomrule
  \end{tabular}
\end{table}

\subsection{Scenario}
\label{subsec:scenarios}

The selected study design aims to provide a plausible scenario, in which two robots interact and one explains its actions. The setting is intended to be natural and simple to understand, while simultaneously difficult to analyze to obscure the objective of the experiment.
To this end, we chose a game scenario, in which two robots play a board game and one explains its moves. The goal of the participant is to bet on the winner of each game, after observing a game-play snippet. Prior work showed that the use of a game scenario can increase participant engangement~\cite{BOYLE2012771}.

According to the aforementioned requirements, the two-player board game Ultimate Tic-Tac-Toe (UTTT) was selected. UTTT is a relatively unknown game with game rules that are quick to understand, but the game is challenging to play due to its large state space~\cite{bertholon2020most, lakhmani2020game}. 
The game board consists of nine nested Tic-tac-toes that compose a larger 3x3 board.
Each smaller board is called a local board and the player has to win three local boards to form a line on the global board, just like in regular Tic-tac-toe.
Thus, it is non-trivial for the participants to analyze the robots' game-play.

\begin{figure}[bp]
    \centering
    \includegraphics[width=0.48\textwidth]{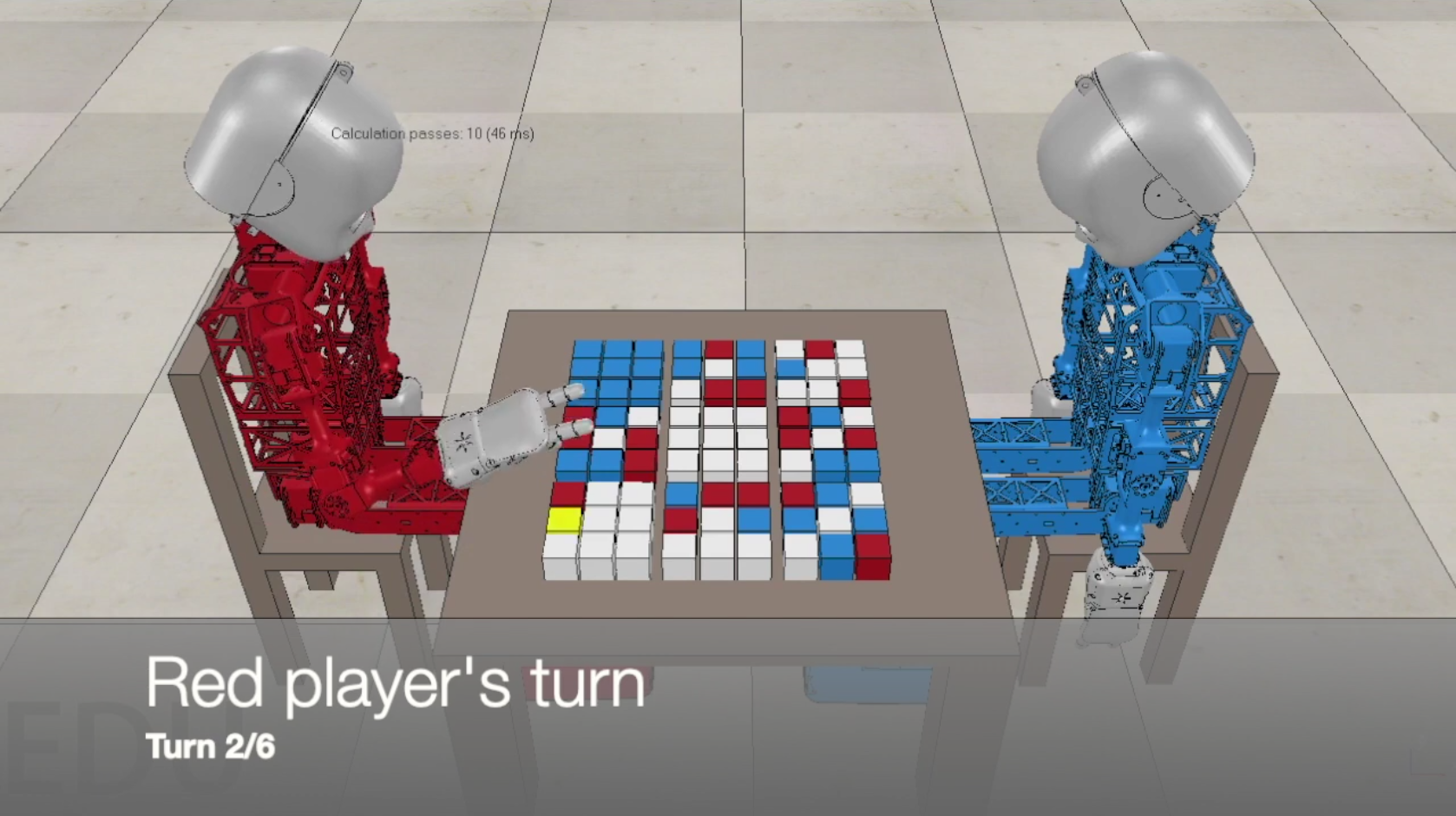}
    \caption{Experiment setup: Two humanoid robots play a game of UTTT against each other in CoppeliaSim. In addition to voice output and robot movement, a video overlay indicates the robot's turn.}
    \label{fig:game-scene}
\end{figure}

To examine the influence of XAI, we use two humanoid robots with different behaviors. Both robots have the same capability to win, as they use the same reinforcement learning algorithm (see Section~\ref{subsubsec:RL}) to play UTTT.
Each turn, the respective robot provides a verbal cue in addition to its game-play.
Both robots can play UTTT against each other autonomously, and simulation software was used to record videos of complete games.
Figure~\ref{fig:game-scene} shows both robots in the simulation environment, sitting at a table across from each other with the UTTT board game in front of them.

\normalcolor

In the study, participants are shown short sequences of pre-recorded games without knowledge about the robots' capabilities and the outcome of the game.
Each short game-play video shows a sequence of three moves by each robot, in which both robots comment on their moves via speech output. While one robot merely announces its moves, the robot using XAI provides an elaboration for the selected move. To actively involve the participants, they have to bet on the robot they believe to win the game. The video snippets were kept short to increase the difficulty of judging the winner from the presented game-play alone, under the assumption that participants would rely more on the robots' utterences to judge their capability.
    
During the pilot study, participants reported difficulties distinguishing the robots in the questionnaire assessment. As a result, the robots were colored differently and were given clearly distinguishable voices.

\subsection{Materials, Software \& Implementation}
\label{subsec:materials}

To enable two humanoid robots to play against each other and conduct the study, a variety of components are required.
The center of the experiment are two Neuro-Inspired Companions (NICOs)~\cite{kerzel2017nico}, which are child-sized humanoid robots. 
The modules that allow these two robots to play UTTT in the simulation software are implemented in the Robot Operating System (ROS)~\cite{ros}.
For simulation and game-play recording, the robot simulation software CoppeliaSim~\cite{coppeliaSim} is used. The voices for both robots are created using
Google Cloud Text-To-Speech\footnote{\url{https://cloud.google.com/text-to-speech}}.
The experiment was conducted using LimeSurvey~\cite{LimeSurvey}. It was used to provide the participants with the game-play videos and assess the questionnaires and bets on the robots.

\subsubsection{Reinforcement Learning}
\label{subsubsec:RL}

To present the study participants with non-trivial and unscripted game-play, a reinforcement learning agent was trained to play the game of UTTT.
The reinforcement learning agent's neural network architecture is based on the model presented by Mnih et al.~\cite{mnih2013playing}, where it is originally used to play Atari games. In contrast to Atari games where each pixel has to be processed, in UTTT only the current board state, estimated by a computer vision model, serves as the agent's input, and the agent returns a probability distribution over the board for the best move.

The agent was trained for $80.000$ games against different types of simple strategy agents and itself. Initially, the agent was bootstrapped with 20.000 games against a strategy that selects a random move unless it can win a local board. Afterward, the agent was trained with 40.000 games against a strategy that does a random move unless it can win a local board or block the opponent from winning a local board. Finally, the agent was trained for 20.000 games against a random mix of the previous strategy (30\%) and self-play (70\%).

\subsubsection{XAI-Algorithm} 
\label{subsubsec:XAIalgorithm}

\label{subsec:expl_types}

For generating explanations of the reinforcement learning agent's moves, a post-hoc approach is utilized.
The XAI-algorithm processes the board and the agent's next move using a set of predefined rules to justify the move. The derived explanations are human justifications of the robot's actions.
Overall, 12 rules were implemented in the XAI-algorithm, which are applied to generate the explanation. As an example of a generated explanation, consider the following scenario: The agent should not play in fields such that his opponent will get a chance to win within his next move. The agent's immediate goal is to force the opponent to play on a different local board. In this case, the explanation will be: "I did not play in row $x$ and column $y$ because my opponent could win the game if I send him to the corresponding local board." A second situation can be that the agent blocks the opponent in a local board, where he already has placed two marks in a row. Given this event, the generated explanation is "I have blocked my opponent here, so he won't be able to win this local board". 

\subsection{Measurements}
\label{subsec:experimentmeasurements}

For the evaluation of the empirical study, the Godspeed questionnaire, perceived competence ratings, and bets on the winning robot are collected.

The Godspeed questionnaire measures the participant's overall perception of the robot. It measures the dimensions of anthropomorphism, animacy, likeability, perceived intelligence, and perceived safety~\cite{bartneckHRIMetrics2009}.

The Perceived Competence Scale (PCS)~\cite{williams1998supporting, williams1996internalization} is applied to measure the robot's competence as perceived by the participant. It is a 4-item questionnaire that was initially used in the Self-Determination Theory~\cite{Ryan2000}.
The questions are adjusted to the subject of the study by converting the perspective of the questions from \enquote{me} to \enquote{robot}.
The first dimension considers the participant's belief that the robot will \enquote{make the right move}, whereas the second dimension assesses their belief of the robot to \enquote{win a game}.

As a direct measure of the robots' perceived competence, the participants are requested to place a bet on the robot they believe to win the game. These bets are assessed after each game-play video.

\subsection{Study Design and Procedure}
\label{subsec:procedure}

The study is constructed as a between- and within-subjects design to investigate how the robot's explanation for its game-play impacts
the participant's perception of the robot.
Specifically, two test groups and one control group are utilized.
In the test group condition, either the red or blue robot explains its moves during the game.
This is in contrast to the control group, where neither robot utilizes XAI and instead only announces its moves.
The experiment conditions can be summarized as follows:

\begin{itemize}
    \item \textbf{XAI red:} The red robot (on the left) explains the moves, while the blue robot only announces its moves.
    \item \textbf{XAI blue:} The blue robot (on the right) explains the moves, while the red robot only announces its moves.
    \item \textbf{Control:} Both robots only announce their moves.
\end{itemize}

For recruitment, participants are provided with a link to the survey and are randomly assigned via LimeSurvey to one of the three conditions (XAI red, XAI blue, control).
The study begins with a general introduction to the experimental procedure. 
Afterward, the participants receive detailed introduction videos for the rules of UTTT and the experiment setup. Then, they are provided with an example game-play snippet and an introduction to the betting procedure. 

\begin{figure}[htbp]
    \centering
    \includegraphics[width=0.48\textwidth]{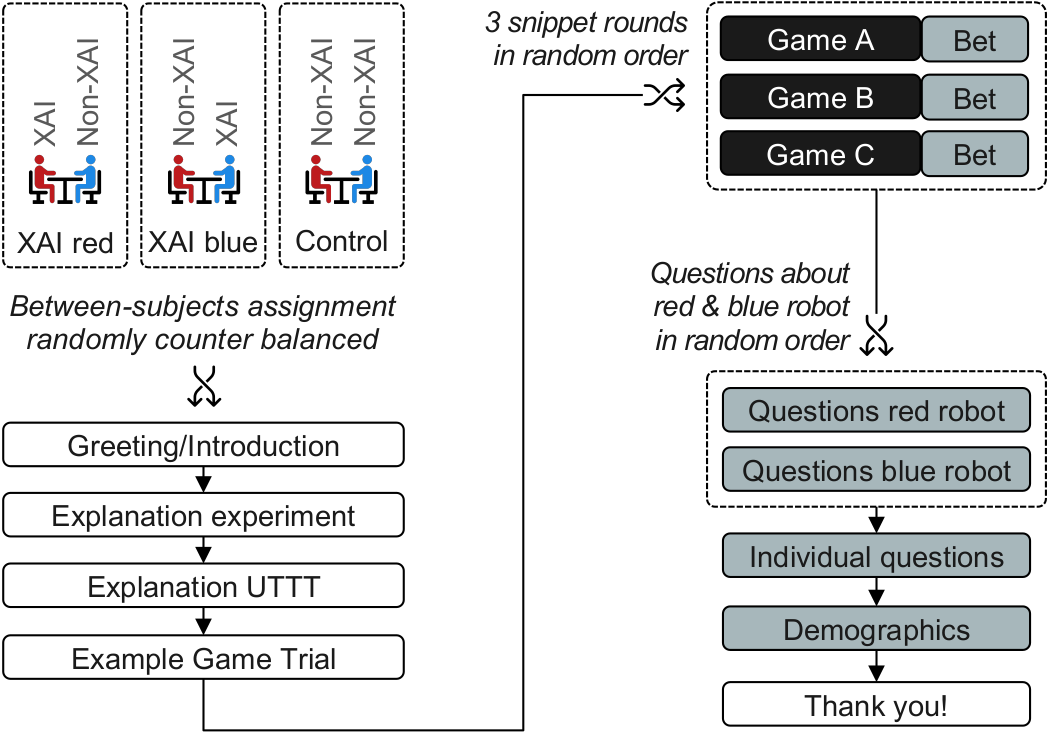}
    \caption{Flow diagram of the experiment procedure}
    \label{fig:procedure}
\end{figure}

After acknowledging to have understood the study procedure, the participants watch three game-play videos of their respective experiment condition in random order. 
Directly after each video, they place their bet on the robot they believe to win the game.
After the experiment, the Godspeed questionnaire, perceived competence questionnaire, and demographics are assessed, where the order of which robot is evaluated first was randomized. 
An overview of the experiment procedure is illustrated in Figure~\ref{fig:procedure}.

\label{methods}
\section{Results}
\label{sec:results}





For measuring the internal consistency of the Godspeed and perceived competence questionnaire, we calculate the Cronbach's Alpha~\cite{cronbach1951coefficient} for each dimension.
Each dimension of both questionnaires shows good internal consistency, except for the safety measure of the Godspeed questionnaire. The dimension of safety only consists of three items and the Cronbach's Alpha indicates an adequate consistency~\cite{Taber2018}. An overview of the estimated Cronbach's Alphas is shown in Table~\ref{table:alphas}.

\begin{table}[htbp]
  \centering
  \caption{Cronbach's Alphas of the questionnaires}
  \label{table:alphas}
  \begin{tabular}{l r r}
      \toprule
      \textbf{Dimension}& \textbf{Godspeed} & \textbf{Godspeed}\\
      & \textbf{Red Robot} & \textbf{Blue Robot}\\

      \midrule
      Anthropomorphism & 0.85 & 0.86\\
      Animacy & 0.90 & 0.89 \\
    Likeability & 0.88&0.89 \\
    Intelligence &  0.85&0.87 \\
    Safety & 0.69 &0.65 \\
      \midrule
      \textbf{Dimension}&  \textbf{Perc. Competence} & \textbf{Perc. Competence}\\
     &  \textbf{Red Robot} & \textbf{Blue Robot}\\

          \midrule
      Right move &0.87 & 0.90 \\
      Ability to win &0.90 &0.88\\      
    \bottomrule
  \end{tabular}
\end{table}



\begin{figure}[htbp]
    \centering
    \includegraphics[width=0.49\textwidth]{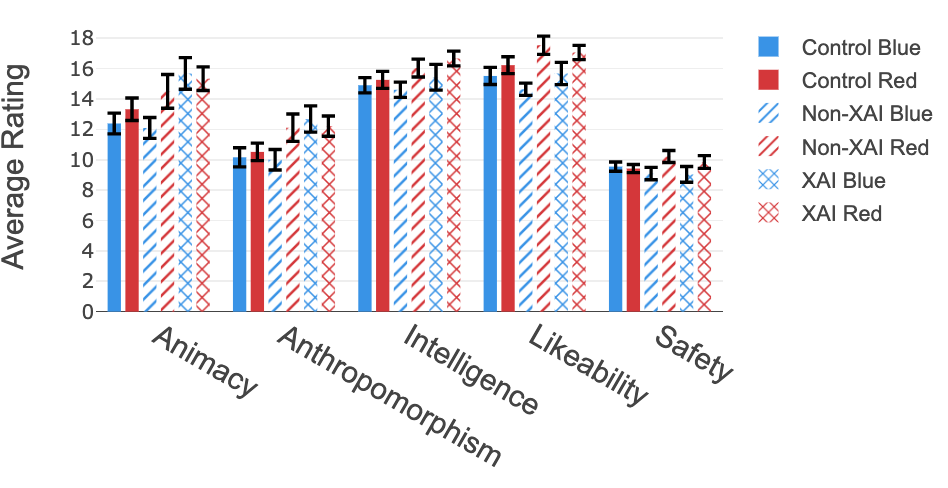}
    \caption{Godspeed dimensions of the individual robots}
    \label{fig:godspeed}
\end{figure}

For the evaluation of the study data, the dimensions of the questionnaires and robots for each experiment condition (control (31 participants), XAI red (33 participants), XAI blue (28 participants)) are separated. This results in the evaluation of six different robots.
An overview of the Godspeed dimensions with estimated means and standard errors is shown in Figure~\ref{fig:godspeed}.

The individual dimensions of the Godspeed questionnaire are evaluated independently utilizing the Kruskal-Wallis test~\cite{kruskal1952use}. In case of significant differences among the robots, we utilize the pairwise Wilcoxon rank sum test as post-hoc analysis to estimate the differences among the robots~\cite{ruxton2006unequal}.

The Kruskal-Wallis test suggests a difference in the anthropomorphism for the different robots ($p$ = 0.038). However, the pairwise Wilcoxon test does not reveal any significant difference. To infer if there is a difference in anthropomorphism between the XAI and non-XAI robot, the measures of anthropomorphism for both robots in the XAI experiment groups are combined. This leads to a sample of the non-XAI robots ($M$ = 10.97, $SE$ = 0.56) and XAI robots ($M$ = 12.42, $SE$ = 0.53). A Wilcoxon test reveals a significant difference between the XAI and non-XAI robot ($p$ = 0.043). 

For animacy, a significant difference is indicated by the Kruskal-Wallis test ($p$ = 0.008). A pairwise Wilcoxon test suggests a significant difference
for the blue XAI robot ($M$ = 15.68, $SE$ = 1.04) and red XAI robot ($M$ = 15.33, $SE$= 0.78) to the blue control robot ($M$ = 12.39, $SE$ = 0.68) and blue non-XAI robot ($M$ = 12.09, $SE$ = 0.69) (Table~\ref{table:Animacy}).

\begin{table}[htbp]
  \centering
  \caption{Pairwise Wilcoxon test for animacy}
  \resizebox{\columnwidth}{!}{%
  \label{table:Animacy}
  \begin{tabular}{l r r r r r}
      \toprule
        & \textbf{Blue Robot} & \textbf{Blue Robot}& \textbf{Blue Robot}& \textbf{Red Robot}& \textbf{Red Robot}\\
       & \textbf{Control} & \textbf{Non-XAI}& \textbf{XAI}& \textbf{Control}  & \textbf{Non-XAI}  \\
      \midrule
Blue Non-XAI& 0.687  &  ---     &  ---     &---     &   ---  \\     
Blue XAI  &  \textbf{0.027} &   \textbf{0.027} &  ---     &---     &   ---    \\  
Red Control&  0.415 &   0.415 &  0.191 &---     &   ---    \\  
Red Non-XAI&  0.415 &   0.314 &  0.415 &0.687 &   ---    \\  
Red XAI& \textbf{0.027} &   \textbf{0.027} &  0.822 &0.191 &   0.415\\  
    \bottomrule
  \end{tabular}
} 
\end{table}





Likewise, for the dimension of likeability, the Kruskal-Wallis test shows a difference in the likeability among the robots ($p$ = 0.003). The post-hoc
pairwise Wilcoxon test (Table~\ref{table:Likeability}) reveals a significant difference among the likeability of the red non-XAI robot ($M$ = 17.54, $SE$ = 0.60) and red XAI robot (M = 17.06, $SE$ = 0.47) to the blue non-XAI robot ($M$ = 14.64, $SE$ = 0.40). Neither the Kruskal-Wallis test for the intelligence ($p$ = 0.102) nor the safety ($p$ = 0.243) indicate a significant difference among the robots.

\begin{table}[htbp]
  \centering
  \caption{Pairwise Wilcoxon test for likeability}
  \label{table:Likeability}
    \resizebox{\columnwidth}{!}{%
  \begin{tabular}{l r r r r r}
      \toprule
        & \textbf{Blue Robot} & \textbf{Blue Robot}& \textbf{Blue Robot}& \textbf{Red Robot}& \textbf{Red Robot}\\
        & \textbf{Control} & \textbf{Non-XAI}& \textbf{XAI}& \textbf{Control}  & \textbf{Non-XAI}  \\
      \midrule
Blue Non-XAI& 0.284  &  ---     &  ---     &---     &   ---  \\     
Blue XAI&  0.799  & 0.276  &  ---     &---     &   ---    \\  
Red Control&  0.603 &  0.134 & 0.763 &---     &   ---    \\  
Red Non-XAI&  0.086  & \textbf{0.004} & 0.156& 0.203  &   ---    \\  
Red XAI&  0.134  & \textbf{0.008} & 0.216 &0.324  & 0.777 \\  
    \bottomrule
  \end{tabular}
  }
\end{table}



\begin{figure}[htbp]
    \centering
    \includegraphics[width=0.49\textwidth]{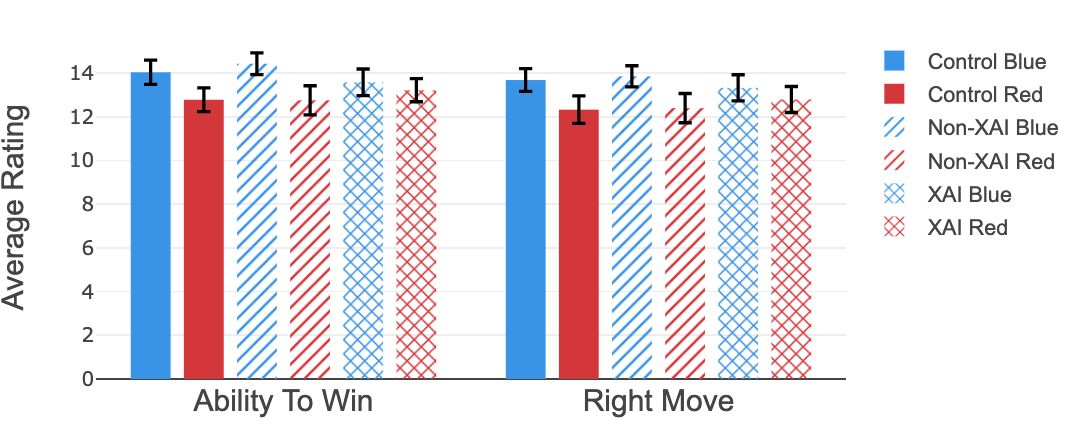}
    \caption{Perceived competence dimensions of the individual robots}
    \label{fig:pcs}
\end{figure}

The perceived competence questionnaire is evaluated identically to the previous procedure, and an illustration of the results of that questionnaire is shown in Figure~\ref{fig:pcs}.
 
The Kruskal-Wallis test for the right move does not suggest a difference among the robots ($p$ = 0.225).
Correspondingly, there appears to be no difference in the robots' ability to win the game ($p$ = 0.205).

The participants' betting behavior is a direct measure of the robots' ability to win the game. For the evaluation, the bets among the three games are aggregated for the different experiment conditions.

\begin{table}[htbp]
  \centering
  \caption{Binomial test for the betting behavior among the experiment conditions}
  \label{table:bet}
  \begin{tabular}{l r r r}
      \toprule
        \textbf{Condition} & \textbf{Bets Blue Robot} & \textbf{Bets Red Robot}& \textbf{p-value}\\
      \midrule
Control& 36  &  57  &  \textbf{0.038} \\     
XAI Blue&  37  &  47   & 0.326 \\  
XAI Red& 46 &  53 &  0.547 \\  
    \bottomrule
  \end{tabular}
\end{table}

\begin{figure}[htbp]
    \centering
    \includegraphics[width=0.35\textwidth]{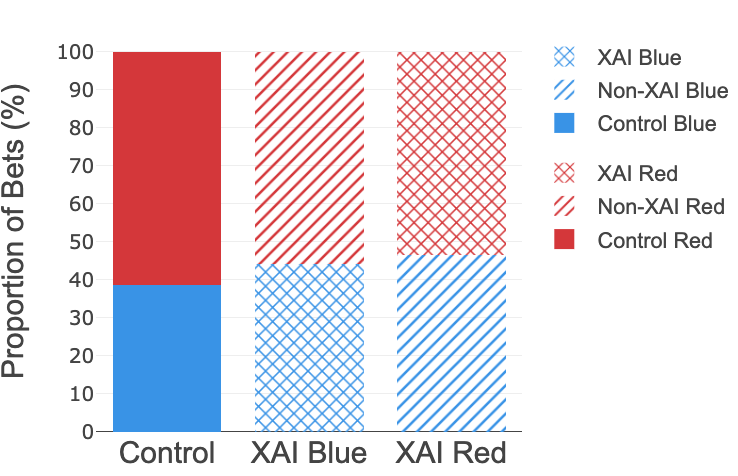}
    \caption{Betting behavior among the experiment conditions}
    \label{fig:bet}
\end{figure}

For each experiment condition, a binomial test is used to estimate if either the red or blue robot was favored during the experiment.
The findings are summarized in Table~\ref{table:bet} and the betting behavior is illustrated in Figure~\ref{fig:bet}.
The results show that in the control group, the participants significantly preferred to bet on the red robot despite both robots behaving identically.

\label{results}
\section{Discussion}
\label{sec:discussion}


The evaluation of the Godspeed questionnaire shows that the XAI robots, which explain their moves to the participant, are perceived as more human-like and lifelike than their non-explaining counterpart. 
Specifically, the behavior of explaining the reasoning behind a move in a board game was perceived as more human-like. Analogously, a significant increase of animacy ratings was observed.
This is not surprising, since the XAI robot was communicating more than the non-XAI robot that only announced the moves during the experiment.

In contrast to the increase in these dimensions, the XAI robot was not perceived as more intelligent, likeable or safer than its counterpart.
The explanation of the robot could have not been precise or convincing enough to affect the perception of the participants, in spite of the majority not having prior experience with the game UTTT. Furthermore, the provided game-play might have shown too few moves to convince the participants of the robot's capability. Instead of relying on the robot's utterences, the participants could have judged the game-play capability on the basis of single moves. In this case, participants would decide on the assumed better player quickly, without correcting their opinion after watching more game-play.
The similarity in safety might arise due to both robots being identical except for one explaining the move, or due to the study being conducted online.

Surprisingly, the perceived competence questionnaire did not reveal a difference in competence between the XAI and non-XAI robots. 
Despite the XAI robot being perceived as more human-like and lively, the perception of the robot's competence was unaffected.
This finding suggests that instead of a relationship among perceived competence and the anthropomorphism or animacy of the robot, there might be a link to perceived intelligence that was similarly unaffected by the XAI robot. 
A reason for not observing a difference in perceived competence could be that the effect of the manipulation by the robot explaining its move was too small. 
Alternatively, the participants might have analyzed only the board-state and game-play to estimate which of the robots will win the game to place their bet. This could have led to the suspicion that both robots have equal capabilities, which could result in a diminished effect of the manipulation.

After each short game-play of the robots, the participants had to decide which robot would win the game.
This betting behavior is a direct measure of confidence in the robot's ability. However, the results show that the XAI robot was not preferably selected as the winner.  
For the control group, the participants favored the red robot despite both exhibiting the same behavior.
This might have had an effect on the study, since it appears that overall the participants favored the red robot over the blue robot.

Preferring the red robot could be a result of assuming the gender of the red robot as female and the blue robot as male. While both robots had a female voice according to Google's Text-To-Speech service, the blue robot's voice was slightly lower. In conjunction with the blue color of the robot, this might have affected the participants' perception of the robot. For example, it has been shown that gendering a robot's voice affects the perception of the robot in terms of stereotypes, preference, and trust~\cite{Nomura2017,Gallimore2019}. Other potential sources for this bias are a general preference of one colour over the other (regardless of gender biases), the type of moves carried out by the robot in the snippets, as well as the side that the robot is sitting on.

The preference of the red robot is also reflected in the assessment of likeability, where the red non-XAI robot was perceived as significantly more likeable than the blue non-XAI robot. Remarkably, providing explanations seems to cancel this effect, leading the blue XAI robot to not be perceived as significantly less likeable than any of the red robots.

\section{Limitations}

The results demonstrate that explanations for actions influence the perception of the robot, however, there are several limitations that need to be addressed in future work.

Since UTTT does not allow symmetric game-play, both robots have to make different moves. Therefore, the selected game sequence might affect the participant's betting behavior and perceived competence. A scenario that allows identical moves might be preferred for the evaluation of the influence of XAI in human-robot interaction.

Further, the study was conducted online, which allows for little control over the engagement of the participants. Participants could have been distracted during the study, consciously evaluated the objective of the study, or might have been confused by the wording in the study without the possibility to ask for clarification. Finally, a higher number of participants for each experiment condition could permit a more accurate estimation of the difference in perception between the robots.
\label{discussion}
\section{Conclusion}
\label{sec:conclusion}

In this study, a human-robot interaction scenario was designed that explores the effect of robots that utilize XAI to explain their actions.
Two robots played the board game UTTT, in which one robot explained the reasoning behind its moves, while the other robots simply announced the next move.
The study was conducted online by presenting three short sequences of game-play. After each sequence, the participants had to bet on the winning robot. The experiment concluded with the assessment of several traits each participant assigned to the robots.

In our findings, we could not show that a robot that explains the motivation behind its actions increases the perceived competence. 
On the one hand, humans might still have limited trust in a robot's ability to perform a specific task despite their provided explanation. On the other hand, to manifest the concept of perceived competence in the participants mind might require more time and evidence of the robot's capability than was provided in the short sequences of game-play.

However, a robot that explained its moves was perceived as more human-like and lively than a robot that only announced the moves. This demonstrates that robots that provide reasoning about their actions influence human perception. Such methods could be used to increase trust in robots, especially since robots and artificial intelligence are often perceived as black boxes. The observed effect might be more pronounced in a human-robot cooperative task.

All the software components of this study are implemented in robot simulation software, which allows the NICO robot to play UTTT autonomously and react to a player's moves.
In future work, a human could play against the robot and the effects on the perception for an XAI and non-XAI opponent could be evaluated. This study illustrates the potential and effects of XAI in HRI, and demonstrates that a robot explaining its behavior can be perceived as more lively and human-like.

\label{conclusion}

\bibliography{bib}
\bibliographystyle{ieeetr}

\appendix

\subsection{Adjusted Perceived Competence Scale} \label{Appendix:adjustedPCSitems}

\subsubsection{"Make the right move"}
\begin{itemize}
    \item I feel confident in the $<blue | red>$ robot's ability to make the right move.					
    \item The $<blue | red>$ robot is capable of making the right move. (The $<blue | red>$ robot has the possibility of doing this in the future.)	
    \item The $<blue | red>$ robot is able to make the right move. (The $<blue | red>$ robot can currently do that.)				\item I feel that the $<blue | red>$ robot is able to meet the challenge of choosing the right move.
\end{itemize}

\subsubsection{"Win a game"}
\begin{itemize}
    \item I feel confident in the $<blue | red>$ robot's ability to win a game.					
    \item The $<blue | red>$ robot is capable of winning a game. (The $<blue | red>$ robot has the possibility of doing this in the future.)
    \item The $<blue | red>$ robot is able to win a game. (The $<blue | red>$ robot can currently do that.)	
    \item I feel that the $<blue | red>$ robot is able to meet the challenge of winning the game.
\end{itemize}
\label{appendix}

\end{document}